# Time Synchronized State Estimation for Incompletely Observed Distribution Systems Using Deep Learning Considering Realistic Measurement Noise

B. Azimian, *Student Member, IEEE*, R. Sen Biswas, *Student Member, IEEE*, A. Pal, *Senior Member, IEEE*, and Lang Tong, *Fellow, IEEE*

*Abstract*— Time-synchronized state estimation is a challenge for distribution systems because of limited real-time observability. This paper addresses this challenge by formulating a deep learning (DL)-based approach to perform unbalanced three-phase distribution system state estimation (DSSE). Initially, a data-driven approach for judicious measurement selection to facilitate reliable state estimation is provided. Then, a deep neural network (DNN) is trained to perform DSSE for systems that are incompletely observed by synchrophasor measurement devices (SMDs). Robustness of the proposed methodology is demonstrated by considering realistic measurement error models for SMDs. A comparative study of the DNN-based DSSE with classical linear state estimation indicates that the DL-based approach gives better accuracy with a significantly smaller number of SMDs.

*Index Terms*— Deep neural network (DNN), Gaussian mixture model (GMM), State estimation, Synchrophasor measurements.

## I. Introduction

Real-time monitoring and control of distribution networks was traditionally felt to be unnecessary because it had unidirectional power flows, radial configuration, and predictable load patterns. However, the fast growth of behind-the-meter (BTM) generation, particularly solar photovoltaic (PV), electric vehicles, and even storage, is transitioning the distribution system from a passive load-serving entity to an active market-ready entity, whose reliable and secure operation necessitates real-time situational awareness [1]-[2]. Synchrophasor measurement devices (SMDs) were introduced in the distribution system to provide fast (sub-second) situational awareness by enabling time-synchronized state estimation [3]; however, their numbers are not large enough to provide an independent estimate of the system state. Moreover, the assumption of Gaussian noise in synchrophasor measurements has been disproved recently [4].

At the same time, modern distribution systems are being equipped with advanced metering infrastructure in the form of smart meters. By the end of 2018, 86.8 million smart meters were installed in the U.S [5]. As such, prior research has combined smart meter data with SMD data for facilitating distribution system state estimation (DSSE) [6]. However, smart meters measure energy consumption from 15 minute to hourly time intervals and report their readings after a few hours or even a few days; these two aspects make smart meter data unsuitable for real-time state estimation. Moreover, smart meter readings are not time-synchronized, which makes their direct integration with SMD measurements a statistical challenge.

Machine learning, in general, and neural networks, in particular, has been extensively used to perform state estimation in distribution systems[7]-[9]. In [7], a shallow physics-aware neural network was used to implement DSSE. However, the approach did not consider the different time resolutions of smart meters, SMDs, and supervisory control and data acquisition (SCADA) measurements. In [8], an artificial neural network framework was used for three-phase unbalanced DSSE. The approach split the grid into multiple areas to improve DSSE accuracy as well as training time. However, smart meter measurements were not considered in the analysis (only micro-PMU measurements were used) and loads were varied by a Gaussian distribution which might not correctly represent actual system behavior. In [9], a Bayesian approach was utilized for DSSE by treating the system states and measurements as random variables. A deep neural network (DNN) was trained to approximate the conditional mean of the joint probability distribution of the system states and measurements. However, the performance of the DNN-based DSSE used in [9] was not validated for unbalanced three-phase distribution systems and non-Gaussian measurement noise. The effect of SMD location on the performance of DNN-based DSSE has also not been investigated in prior work.

This paper overcomes the challenges outlined above by making the following salient contributions:
1. A DNN-based DSSE is developed for *unbalanced three-phase* distribution networks that estimates system states (voltage phasors) in a fast, time-synchronized manner.
2. The robustness of the proposed method is demonstrated by considering *realistic errors* in SMD measurements.
3. A judicious approach for *measurement selection* inside a DNN framework to facilitate reliable fast time-synchronized state estimation is presented.

## II. Distribution System State Estimation (DSSE) using Deep Neural Network (DNN)

### A. Bayesian Approach to DSSE

To circumvent the problem of scarcity of SMDs, a Bayesian approach is implemented for DSSE. In a Bayesian framework,

This work was supported in part by the Advanced Research Projects Agency-Energy (ARPA-E) under award number DE-AR00001858-1631 and by the Power Systems Engineering Research Center (PSERC) Grant T-63.
B. Azimian, R. Sen Biswas, and A. Pal are with the School of Electrical, Computer and Energy Engineering, Arizona State University, Tempe, AZ, 85287, USA (e-mail: bazimian@asu.edu; rsenbisw@asu.edu; apal12@asu.edu).
L. Tong is with the School of Electrical and Computer Engineering, Cornell University, Ithaca, NY 14850, USA (e-mail: lt35@cornell.edu).

both the state, $x$, and the measurement, $z$, are treated as random variables. Similar to [9], we create a minimum mean squared error (MMSE) estimator to directly minimize the estimation error as shown below.

$$\min_{\hat{x}(\cdot)} \mathbb{E}(\| x - \hat{x}(z) \|^2) \implies \hat{x}^*(z) = \mathbb{E}(x|z) \qquad (1)$$

It should be noted that the MMSE estimator directly minimizes the estimation error while classical estimators, such as least squares, minimize the modeling error, which is embedded in the measurement function $h(\cdot)$ that relates the measurements with the states, i.e., $z = h(x) + e$. By circumventing the need for $h(\cdot)$, the observability requirements get bypassed in a Bayesian state estimator. However, in (1), there are two underlying challenges to computing the conditional mean. First, the conditional expectation, which is defined by,

$$\mathbb{E}(x|z) = \int_{-\infty}^{+\infty} x p(x|z) dx \qquad (2)$$

requires the knowledge of joint probability density function (PDF) between $x$ and $z$, denoted by $p(x,z)$. When the number of SMDs are scarce, the PDF between SMD measurements and all the voltage phasors is unknown or impossible to specify, making the direct computation of $\hat{x}^*(z)$ intractable. Secondly, even if the underlying joint PDF is known, finding a closed-form solution for (2) can be difficult. DNN is used in this paper to approximate the MMSE state estimator.

*B. Deep Learning (DL) using Deep Neural Network (DNN)*

According to the universal approximation theorem [10], a neural network is capable of approximating any arbitrary continuous function. As the MMSE estimator is a regression problem with the regressor, $x$, it can be implied that the MMSE estimator in (1) can be approximated by a DNN that can capture the non-linearities and complexities of the distribution system, such as unbalanced loads, presence of single-and-two-phase laterals, varying regulator taps and capacitor banks, and incomplete observability for different operating conditions. Consequently, a Bayesian DSSE can be implemented by training a regression DNN offline, and then using the trained DNN during real-time operation to estimate the states. The inputs to the DNN will be the $z$ obtained from SMDs and the outputs will be the estimated voltage phasors, $\hat{x}^*(z)$.

The proposed DNN structure is shown in Figure 1, where $m$ is the total number of measurements available from SMDs, $n$ is the total number of states, $a$ is the activation function for each neuron, $b$ is the bias for each neuron, and $W$ refers to the weights conveying the output of previous neurons to the neurons of the next layer. For an incompletely observed distribution network, $n$ is greater than $m$. The number of neurons and hidden layers are hyperparameters that must be tuned offline. A rectified linear unit (ReLU) activation function is used for the hidden layers, while a linear activation function is used for the output layer. The loss function is chosen to be the empirical mean-square error which is consistent with the Bayesian approach. Robust DNN training requires large amounts of data. This data is created by running multiple power flows to generate the required measurement-state $(z, x)$ samples. During the offline training process the weights are optimized to minimize the mean squared error using the backpropagation [18] algorithm. In real-time operation, the data from SMDs are fed into the trained feed-forward DNN and the estimated state, $\hat{x}^*$, is obtained.

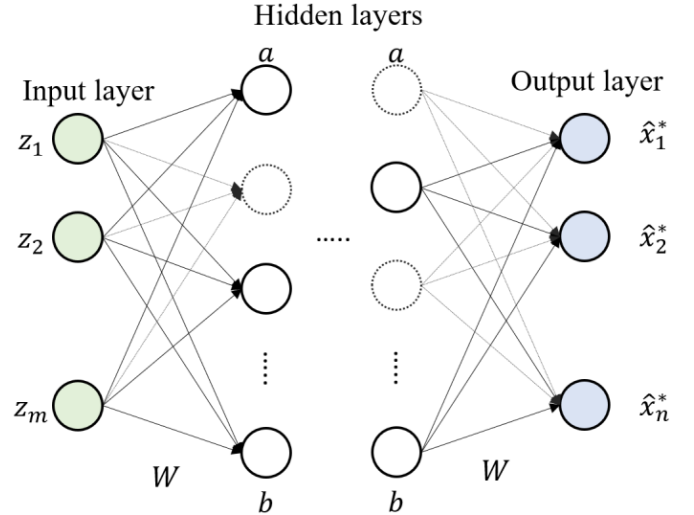

Figure 1: DNN structure for DNN-based DSSE

In the proposed analysis, the unique attributes of distribution networks are considered. For the transmission system, it is common to assume that the three-phase system is balanced. But this assumption is not valid for distribution systems due to the presence of unbalanced loads and single-and-two-phase laterals [11]. An SMD typically has six channels which measure three nodal voltage phasors and three branch current phasors [3]. This provides observability of individual phases at the node equipped with the SMD. Each phasor magnitude and angle are separate features that are fed into the input layer of the DNN. In the output layer, the voltage magnitude and angle of every phase of every node is estimated. Moreover, other distribution network characteristics such as wye-delta loads, phase-order, zero-injection phases (ZIPs), voltage regulators, transformers, and capacitor banks are included in the physical model of the network that is used for creating samples for the DNN training.

*C. Sample Generation and Distribution Learning*

As mentioned in Section I, smart meter measurements become available after a latency of at least a few hours, implying that these measurements cannot be used for real-time DSSE. Moreover, the number of SMDs in a practical distribution system is not large enough to provide an independent state estimate. Thus, the proposed methodology takes advantage of the available historical slow timescale smart meter readings in the offline training process of the DNN. The online stage only needs limited SMDs to carry out DSSE in real-time. In the offline stage, the smart meter energy readings are converted to average power by dividing the energy with the corresponding time interval. Then, the aggregated net injection at the distribution transformer level is calculated by summing up the readings of the smart meters connected to it. The net load at each transformer is treated as a random variable.

Next, a *kernel density estimator* (KDE) is used to learn the distribution of aggregated smart meter readings. Although KDE is suitable for learning the PDF of data samples which do not exactly follow a parametric PDF, it is prone to overfitting which causes loss of generality of the fitted PDF. We modify the KDE by adjusting its bandwidth to achieve the 95% confidence interval limits to ensure that the fitted PDF can effectively represent the load behavior. After the PDF of active power injection is obtained, the reactive power is computed by selecting a power factor from a uniform distribution lying between 0.95 and 1. After the distributions are learned, Monte Carlo (MC) sampling is used to generate active and reactive power injections. Power flow is run for each scenario and all voltage and current phasors are stored for each MC sample.

*D. Two-level Error Model for SMDs*

According to the IEEE standard [12], SMDs should meet the total vector error (TVE) requirement of 1%. At the same time, it has also been widely assumed that the errors in SMD data follow a Gaussian [19]. However, SMDs are connected to the grid through an instrumentation channel consisting of instrument transformers, cables, and burden. These components will not only cause the total measurement error to go beyond the 1% TVE limit, but also change the shape of the error distribution; e.g., from Gaussian to a 3-component Gaussian mixture model (GMM) [4]. Note that the instrument transformer error alone for voltage magnitudes, voltage angles, current magnitudes, and current angles can be as high as ±1.2%, ±1°, ±2.4% and ±2°, respectively [13].

To account for these practical constraints, a two-level error model is introduced in this paper, as shown in Figure 2. In Figure 2, $V_{grid}$ and $I_{grid}$ are the true voltages and currents generated from power flow analysis for different scenarios. In the first level, the instrumentation channel error, is modeled by a 3-component GMM with the corresponding magnitude and angle errors added to the true voltages and currents of the grid. In the next level, a Gaussian TVE is added on top of the previously obtained erroneous measurements to generate the final voltage and current measurements from SMDs. This two-level error model ensures the generation of realistic SMD data.

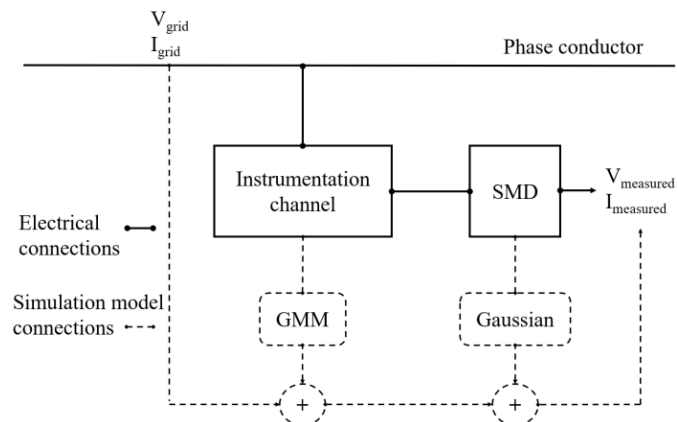

Figure 2: Two level error model for SMDs

*E. Overview of DNN-based DSSE*

Figure 3 provides an overview of the proposed methodology. The model is split into an offline learning stage and a real-time operation stage. In the offline learning stage, historical smart meter data is used to find the PDFs of the power injections at a given node. MC sampling of power injections is done from the best-fit PDF to create different scenarios. Three-phase unbalanced power flow is solved in OpenDSS for the different scenarios. All voltage phasors and corresponding SMD measurements are saved for each scenario. The DNN is trained with the generated measurements. For the real-time operation, the trained DNN is used to perform DSSE using streaming data obtained from the SMDs.

## III. MEASUREMENT SELECTION

DNN-based DSSE is a regression problem, in which all voltages and currents of the distribution network can be considered as potential input features. Hence, measurement selection can be viewed as a *feature selection* problem, whose objective is to find the minimum number of SMDs required to achieve an acceptable DSSE performance. The most common technique to find the best features for a regression problem is to use correlation coefficients [14]. In this paper, we have used Spearman's correlation coefficient for feature selection. It was observed that the presence of transformers and multiple outgoing laterals from the feeder head splits the correlation coefficient matrix into multiple clusters. The number of identified clusters can be a good indicator of the minimum number of required SMDs. It should be noted that adding more SMDs to the same cluster will not necessarily improve DNN performance because (a) features in the same cluster do not provide extra information, and (b) it may increase the possibility of overfitting, which may even deteriorate (DNN) performance. Thus, among the nodes that belong to the same cluster, the location that facilitates maximum observability was found to be a suitable candidate for installation of the SMD. In this context, an index called the phase observability index (POI), based on the bus observability index (BOI) of [15], and defined as the *total number of phase voltages that can be observed from a given location*, was used to screen out the location for SMD installation.

## IV. SIMULATION RESULTS

*A. Simulation Settings*

*1) Distribution system setup*

Simulations are performed on the IEEE 34-node system (System S1) and a 240-node distribution network located in the Midwest U.S. (System S2). For System S1, the loads are varied by 50% based on a Gaussian distribution. System S2 belongs to a municipal utility that has smart meters installed at all customer premises [16]. It has all the characteristics of a modern distribution network: underground and overhead lines, capacitors and regulators, single, double, and three-phase laterals and loads. One-year worth of smart meter readings is also available for this system. PDFs (using modified KDE) were fit to the historical hourly smart meter data, while ensuring that there was no overfitting. Distribution network models of Systems S1 and S2 are available in OpenDSS [17].

*2) Neural network setup*

Hyperparameter tuning is the most crucial part of DNN training. The hyperparameter information for the regression DNN is summarized in Table I. Keras v. 2.2.4 with Tensorflow v.2.1.0 as the backend was used in Python v.3.7.7 to carry out the training. All simulations were performed on a computer with 64.0 GB RAM and Intel Core i7-8700k CPU @3.70GHz

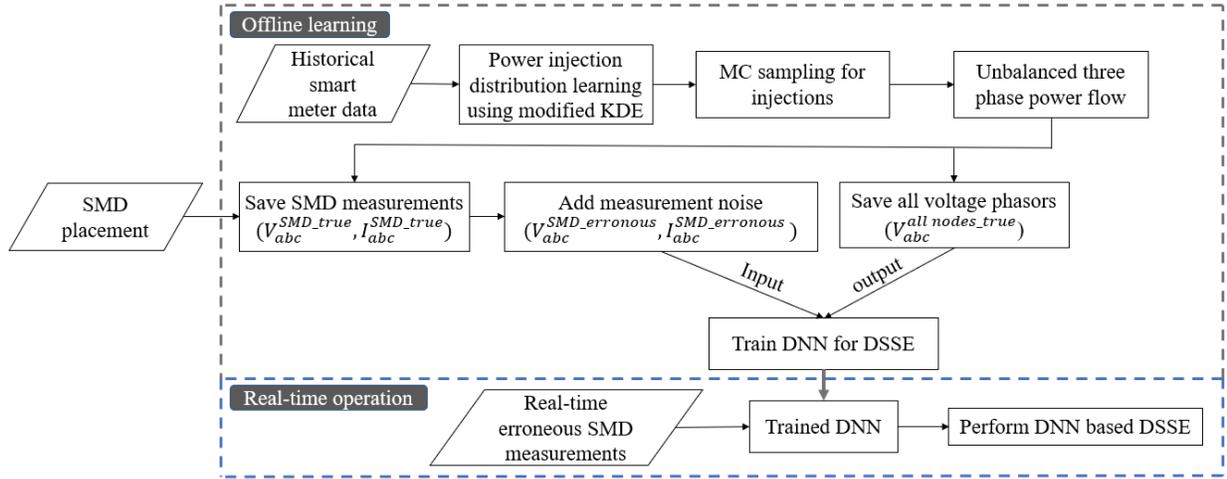

Figure 3: Overview of the proposed DNN-based DSSE

**Table I: Hyperparameters for DNN-based DSSE**

| Hyperparameter | DNN-based DSSE |
|---|---|
| No. of neurons in input layer | 2×No. of measured phasors by all SMDs |
| No. of neurons in each hidden layer | 500 |
| No. of hidden layers | 5 |
| No. of output neurons | No. of states |
| Hidden layer activation function | ReLU |
| Output layer activation function | Linear |
| Initializer method | He normal |
| Optimizer | ADAM |
| No. of epochs | 200 |
| No. of samples | 12,500 |
| Learning rate (lr) | 0.1 with reduce lr on Plateau |
| Regularization | 30% Dropout |
| Loss function | Mean squared error |

*B. IEEE 34-node system (System S1)*

Figure 4 shows the heatmap for the Spearman's correlation coefficient between phase A voltage phasor angles of this system (a similar observation was made for the other phases as well). From the figure, it can be seen that System S1 is composed of two clusters: a smaller cluster comprising nodes 888 and 890, and a larger cluster, containing all the remaining nodes. This leads to the conclusion that two SMDs (one per cluster) is appropriate for DNN-based DSSE for this system. Figure 5 shows the phase mean absolute error (MAE) for DNN-based DSSE when: Case (a) one SMD is placed in the system inside the large cluster at a location (808-812) that has high POI (red dots); Case (b) two SMDs are placed in the system at two locations (808-812 and 830-854) that have high POI but are in the same (large) cluster (black cross); Case (c) two SMDs are placed in the system at two locations (808-812 and 888-890) that have high POI but are in different clusters (blue dots), and case (d) one SMD is placed in the system inside the small cluster at a location (888-890) that has high POI (blue cross). From the figure, it is clear that the MAE decreased considerably when the two SMDs were placed in two different clusters (Case (c)), thereby validating the logic proposed in Section III.

Next, the performance of DNN-based DSSE is compared with the classical linear state estimation (LSE) [20] for System S1. To satisfy LSE's complete observability requirement, this system needed 26 SMDs [11]. It can be observed from Table II that DNN-based DSSE outperforms classical LSE in terms of both phase MAE and magnitude mean absolute percentage error (MAPE) with only two SMDs. It should be noted that for comparing the DNN-based DSSE approach with LSE, the GMM error was removed from the data generation process of LSE and only 1% TVE for SMDs was considered. This is because the least-squares method is guaranteed to give the best possible result in situations where the measurement noise follows a Gaussian distribution. Lastly, it was observed that the trained DNN-based DSSE was able to easily match the output rates of typical SMDs (e.g., 30 samples per second). Therefore, it is considerably fast in comparison to other approaches that rely only on smart meters to do DSSE (such as [21]).

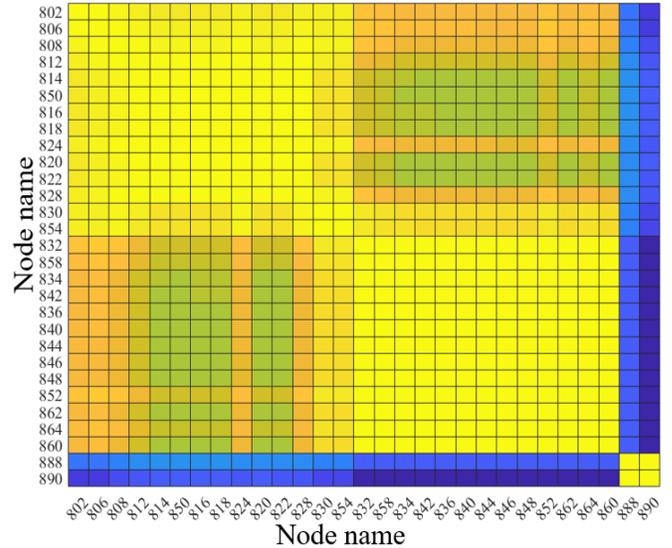

Figure 4: Spearman's correlation coefficient for phase A voltage angles for System S1

*C. 240-node distribution network of Midwest US (System S2)*

A similar approach was implemented for measurement selection for System S2. In the Spearman's correlation coefficient heatmap for this system, three distinct clusters were observed, which corresponded to the three feeders present in System S2 [16]. Therefore, three SMDs were installed at 1001-1002, 2036-2037, and 3075-3076 (one in each feeder) for DNN-based DSSE. Based upon POI, these locations provided high phase observability in the network. Next, the performance of DNN-based DSSE was compared with the classical LSE for this system. The total number of SMDs required for complete observability of System S2 was 113 based on the optimization framework proposed in [11]. It can be observed from Table III that the DNN-based DSSE outperforms classical LSE with three SMDs. Moreover, the accuracy of DNN-based DSSE is practically the same with 1% Gaussian TVE and with the two level-GMM error model. This shows that the DNN-based DSSE is robust against both the error model and the error magnitude.

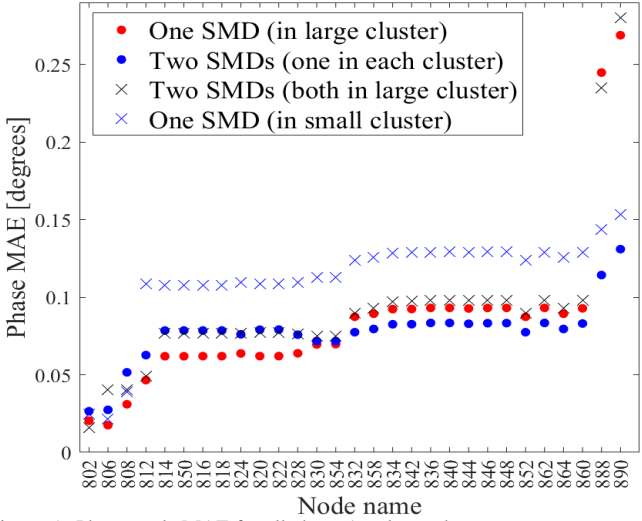

Figure 5: Phase angle MAE for all phase A voltage phasors

**Table II: Comparing the performance of DNN-based DSSE with classical LSE for System S1**

| Method | Error model | Phase MAE [degrees] | Magnitude MAPE [%] | # SMD |
|---|---|---|---|---|
| LSE | 1% Gaussian TVE | 0.14 | 0.25 | 26 |
| DNN-based DSSE | 1% Gaussian TVE | 0.07 | 0.18 | 2 |
| DNN-based DSSE | Two-level GMM | 0.08 | 0.18 | 2 |

**Table III: Comparing the performance of DNN-based DSSE with classical LSE for System S2**

| Method | Error model | Phase MAE [degrees] | Magnitude MAPE [%] | # SMD |
|---|---|---|---|---|
| LSE | 1% Gaussian TVE | 0.14 | 0.25 | 113 |
| DNN-based DSSE | 1% Gaussian TVE | 0.03 | 0.08 | 3 |
| DNN-based DSSE | Two-level GMM | 0.03 | 0.09 | 3 |

## V. CONCLUSION

In this paper, a DNN framework for unbalanced three-phase time-synchronized DSSE is proposed that does not require complete observability of the network by SMDs. A data-driven methodology for measurement selection is initially proposed to enhance the performance of DNN-based DSSE. Finally, the performance of the proposed methodology is validated by comparing it with the classical LSE. The simulation results on the IEEE 34-node distribution feeder and the 240-node Midwest US system show that the proposed method: (1) can achieve better accuracy with a significantly smaller number of SMDs, and (2) is robust against non-Gaussian measurement noise. The ability of the proposed algorithm to provide reliable state estimates with few SMDs in large distribution networks makes it a suitable candidate for enhanced distribution system monitoring and control applications. Investigating the effects of BTM solar PV generation, varying topologies, and uncertain network models on the performance of the proposed DNN-based DSSE as well as more rigorous ways of doing hyperparameter tuning will be explored in the future.